\documentclass[10pt,twocolumn,letterpaper]{article}

\usepackage[pagenumbers]{cvpr} 

\usepackage{graphicx}
\usepackage{amsmath}
\usepackage{amssymb}
\usepackage{booktabs}
\usepackage{float}
\usepackage{titling}
\usepackage{graphics}
\usepackage{xspace}
\usepackage{color}
\usepackage{bm}

\usepackage{siunitx}
\usepackage{balance}
\usepackage{lmodern}
\usepackage{tabularx}
\usepackage{booktabs}
\usepackage{makecell}
\usepackage[hyphens]{url}
\usepackage{amssymb}

\usepackage[pagebackref,breaklinks,colorlinks]{hyperref}

\usepackage[capitalize]{cleveref}
\crefname{section}{Sec.}{Secs.}
\Crefname{section}{Section}{Sections}
\Crefname{table}{Table}{Tables}
\crefname{table}{Tab.}{Tabs.}

\newcommand{\methodname}{GazeNeRF\xspace}
\newcommand{\face}{face only\xspace}
\newcommand{\eyes}{eyes\xspace}
\newcommand{\mlps}{two-stream MLPs\xspace}

\newcolumntype{P}[1]{>{\raggedright\arraybackslash}m{#1}}%
\newcolumntype{C}[1]{>{\centering\arraybackslash}m{#1}}%
\newcolumntype{R}[1]{>{\raggedleft\arraybackslash}m{#1}}%

\usepackage{xcolor}

\usepackage[colorlinks]{hyperref} 

\definecolor{xucongcolor}{rgb}{0.73725, 0.6588, 0.0705} 

\newif\ifshowcomments
\showcommentstrue 

\usepackage[normalem]{ulem}
  
\newif\ifshowchange
\showchangetrue 

\usepackage{enumitem}
\usepackage{paralist}
\usepackage{multicol}

\usepackage[pagebackref,breaklinks,colorlinks]{hyperref}
\begin{document}

\title{\methodname: 3D-Aware Gaze Redirection with Neural Radiance Fields}

\author{Alessandro Ruzzi\textsuperscript{1}\thanks{These two authors contributed equally to this work.} \quad Xiangwei Shi\textsuperscript{2}\footnotemark[1] \quad Xi Wang\textsuperscript{1}  \quad Gengyan Li\textsuperscript{1} \quad Shalini De  Mello\textsuperscript{3} \\ \quad Hyung Jin Chang\textsuperscript{4} \quad Xucong Zhang\textsuperscript{2} \quad Otmar Hilliges\textsuperscript{1}
\\ 
    \normalsize\textsuperscript{1}Department of Computer Science, ETH Zürich \quad 
    \textsuperscript{2}Computer Vision Lab, Delft University of Technology \quad 
    \normalsize\textsuperscript{3}NVIDIA \quad \\
    \normalsize\textsuperscript{4}School of Computer Science, University of Birmingham \quad \\
}

\date{}

\pagestyle{empty}

\maketitle
\pagestyle{empty}
\thispagestyle{empty}

\begin{abstract}
  We propose \methodname, a 3D-aware method for the task of gaze redirection.
Existing gaze redirection methods operate on 2D images and struggle to generate 3D consistent results.
Instead, we build on the intuition that the face region and eyeballs are separate 3D structures that move in a coordinated yet independent fashion. 
Our method leverages recent advancements in conditional image-based neural radiance fields and proposes a two-stream architecture that predicts volumetric features for the face and eye regions separately. 
Rigidly transforming the eye features via a 3D rotation matrix provides fine-grained control over the desired gaze angle.
The final, redirected image is then attained via differentiable volume compositing. 
Our experiments show that this architecture outperforms na\"{i}vely conditioned NeRF baselines as well as previous state-of-the-art 2D gaze redirection methods in terms of redirection accuracy and identity preservation. 
Code and models will be released for research purposes.
\end{abstract}

\section{Introduction}

Gaze redirection is the task of manipulating an input image of a face such that the face in the output image appears to look at a given target direction, without changing the identity or other latent parameters of the subject.
Gaze redirection finds applications in video conferencing \cite{wolf2010eye}, image and movie editing \cite{deepwarp}, human-computer interaction \cite{park2021talking}, and holds the potential to enhance life-likeness of avatars for the metaverse (e.g., \cite{zhou2021pose,gafni2021dynamic}). 
It has furthermore been shown that gaze-redirected images can be used to synthesize training data for downstream tasks such as person-specific gaze estimation~\cite{selflearning,he2019gazeredirection}.

\begin{figure}[t]
    \centering  
    \includegraphics[width=0.46\textwidth]{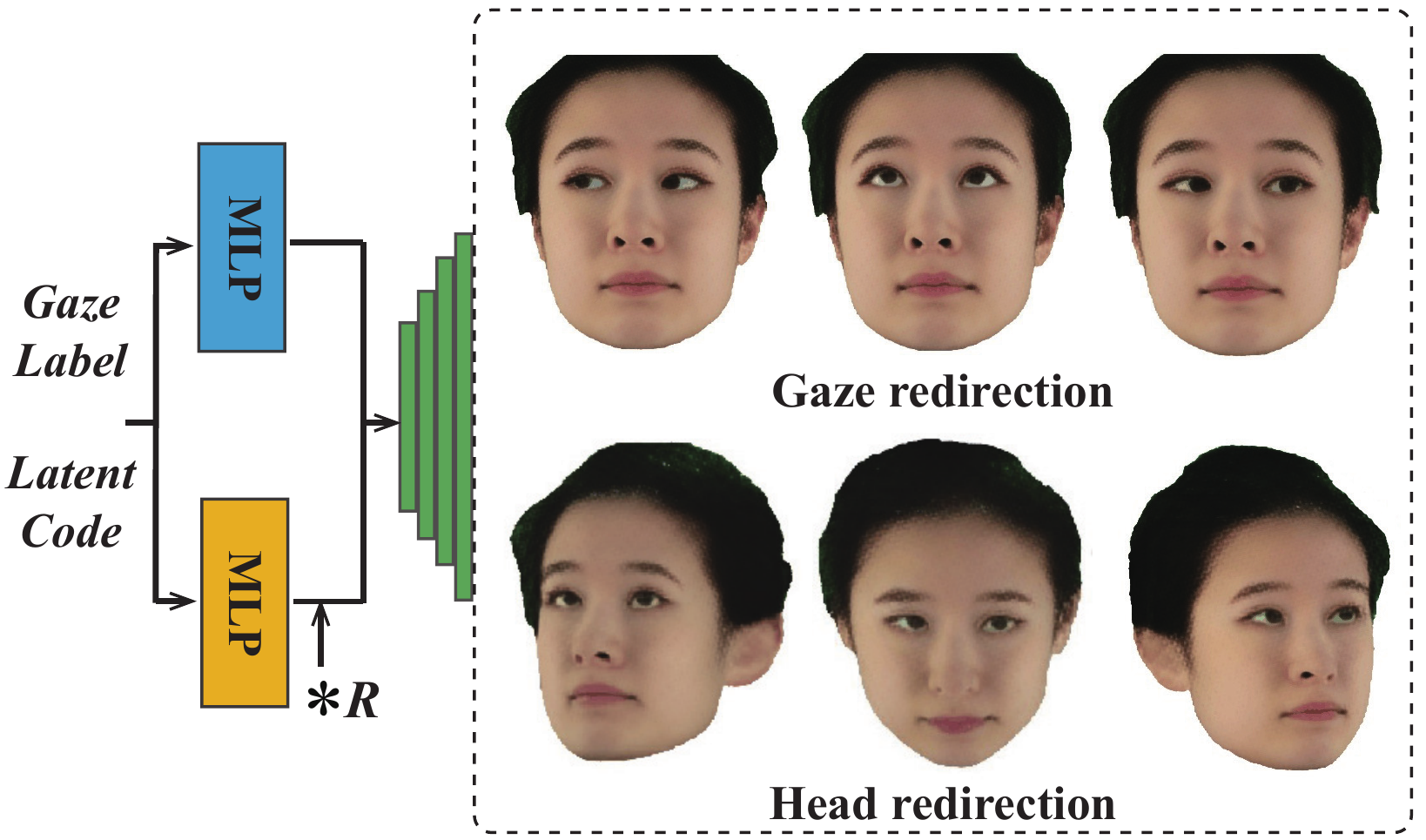}
    \caption{\methodname consists of a NeRF-based two-stream-MLP structure conditioned on a target gaze label to generate the photo-realistic face images. A 3D rotation transformation $\textbf{R}$ is applied on eyes stream of \methodname.}
    \label{fig:teaser}
\end{figure}

Existing gaze redirection methods formulate this task as a 2D image manipulation problem.
Either by warping select pixels of the input image~\cite{deepwarp, zhang18_etra,zhang2022gazeonce,yu2020unsupervised}, or by synthesizing new images via deep generative models such as Generative Adversarial Networks (GANs) \cite{he2019gazeredirection,jindal2021cuda}, encoder-decoder networks \cite{park2019few}, or Variational Autoencoders (VAEs) \cite{selflearning}. Image warping methods can not model large changes due to the inability to generate new pixels. While 2D generative models can produce high-quality images and allow for large gaze direction changes, they do not take the 3D nature of the task into consideration. 
This can lead to spatio-temporal or identity inconsistencies where other latent variables are entangled with the gaze direction. 
Some 2D methods attempt to simulate the eyeball rotation by applying a 3D rotation matrix in latent space~\cite{park2019few, selflearning}. 
However, these injected implicit priors are weak and do not explicitly model the 3D nature of the task.

In this paper, we address these issues by reformulating gaze redirection as a 3D task and propose a novel 3D-aware gaze redirection method \methodname. Our approach leverages recent advances in image-based conditional neural radiance fields~\cite{headnerf} to inherit the ability to generate images of excellent quality. 
The physical face and eyes are not a monolithic 3D structure but are composed of two 3D structures -- the face without eyes that deforms and the eyes only that rotates when we move our eyes. Hence, we model the two structures as separate feature volumes with neural radiance field (NeRF) models.
To this end, our work shares similarities to EyeNeRF~\cite{li2022eyenerf}, but their focus is on high-fidelity rendering and relighting quality, whereas we are concerned with gaze redirection accuracy.

To endow NeRF architectures with 3D-aware gaze redirection capabilities, 
we propose a novel two-stream multilayer-perceptron (MLP) structure that predicts feature maps for the eye-balls (\textit{\eyes}) and the rest of the face region (\textit{\face}) separately (see \cref{fig:teaser}). The features of the \eyes region are transformed via the desired 3D rotation matrix, before compositing the regions via differentiable volume rendering.
With the explicit separation of the eyeballs, \methodname rigidly rotates the 3D features which we show to be beneficial for gaze redirection accuracy. 
To be able to train the model, we propose the feature composition at end of the \mlps and additional training losses to  enhance the functionality of gaze redirection. 

We find that \methodname outperforms previous state-of-the-art methods~\cite{selflearning,headnerf} for gaze redirection on multiple datasets in terms of gaze and head pose redirection accuracy and identity preservation, evidencing the advantage of formulating the task as a 3D-aware problem.
In summary, our contributions are as follows:
\begin{compactitem}
    \item We re-formulate the task of gaze redirection as 3D-aware neural volume rendering.
    \item \methodname learns to disentangle the features of the face and eye regions, which allows for the rigid transformation of the eyeballs to the desired gaze direction. 
    \item State-of-the-art performance in gaze redirection accuracy under identity preservation across different datasets. 
\end{compactitem}
\section{Related Work}

\subsection{Gaze redirection}
Gaze redirection can be done with the graphic model that synthesizes the eye images with different gaze directions and head poses \cite{gazedirector}.
However, these methods are expensive due to the complex appearance modelling such as albedo, diffuse, shading and illuminations, which usually require accurate facial and eye landmarks detection.
Recent gaze redirection methods mainly utilize image warping method~\cite{deepwarp,zhang2022gazeonce,yu2020unsupervised} or generative models~\cite{park2019few,selflearning,he2019gazeredirection} to redirect the gaze and/or rotate the head. 
Image warping estimates warping matrices between source and target images and copies the pixels from the source image to the target image \cite{deepwarp,yu2020unsupervised}.
However, the image warping method cannot generate new pixels out of the input image, which limits its ability for the target gaze label that is far away from the source image gaze direction. 

To overcome this limitation, generative models have been used to synthesize the face or eye images with the target gaze label. He \etal \cite{he2019gazeredirection} introduce a GAN-based approach to generate photo-realistic images with cycle consistency training.
To regularize the generated images, they train a gaze estimator to produce the gaze estimation loss between the generated eye image and the ground truth eye image.
Xia \etal propose controllable gaze redirection methods that explicitly control the gaze with an encoder-decoder structure \cite{xia2020controllable}.
STED \cite{selflearning} proposes a VAE architecture following FAZE \cite{park2019few} with the extension to generate the full-face image instead of the eye patch.
Nonetheless, they require a pair of labeled samples during training.

However, none of the existing methods of gaze redirection is explicitly 3D aware, even though rigid eyeball rotation is inherently a 3D problem. 
STED \cite{selflearning} and FAZE \cite{park2019few} introduce an explicit rotation on the learned latent representations while it is a very weak prior nonetheless.
The rotation matrix is applied to the 2D feature out of an encoder architecture which is mixed with the eyes and the rest of the face. 
Essentially, such rotation operation ignores the 3D nature of rigid eyeball rotation and the deformation of the rest of the face.
To introduce the actual 3D eyeball rotation, EyeNeRF \cite{li2022eyenerf} presents a graphics-based method that fully models the eyeball and the periocular region, yet the focus of EyeNeRF is more on perceptual image quality and photo-realism applications and no result of redirection fidelity is reported in \cite{li2022eyenerf}. 
To train this complex model, EyeNeRF requires a large amount of high-quality data as input, including multiple videos from different camera views accounting for up to 40 minutes.
In contrast, \methodname incorporates the 3D awareness with gaze redirection by applying the rotation matrix to the disentangled eyeball feature maps.

\subsection{Neural Radiance Fields}
Mildenhall \etal~\cite{nerf} propose Neural Radiance Fields to represent a static scene with multi-layer perceptrons. NeRF implicitly learns a 3D-aware continuous function and maps the 3D positions and viewing directions to a density and radiance, which is used for generating novel views with volume rendering. Many following works~\cite{headnerf,niemeyer2021giraffe,park2021nerfies,lazova2022control,jang2021codenerf,zhuang2022mofanerf} focus on controlling the NeRF-based models to represent the dynamic scenes. Hong~\etal~\cite{headnerf} propose HeadNeRF, a NeRF-based model to generate high-fidelity head images by controlling the shape, expression, and albedo of the faces with different illumination conditions. They bring the facial parameters from the 3D morphable model (3DMM) into the NeRF-based model and train the HeadNeRF to generate the dynamic head images conditioned on those learnable latent codes. HeadNeRF can synthesize head images with excellent perceptual quality and add the controllability of facial identity and motion. Some similar works to \cite{headnerf} also generate dynamic faces by controlling the shape, expression, and appearance latent codes of the faces in~\cite{zhuang2022mofanerf}. However, the existing NeRF-based methods of face generation lack gaze control. Different from the previous works, our work focuses on gaze redirection with a NeRF-based model. To control the gaze direction, we train the NeRF-based model conditioned on the gaze label and rigidly rotate the 3D features of the eyes.

\section{Method}
\begin{figure*}[htbp]
    \centering
    \includegraphics[width=0.95\textwidth]{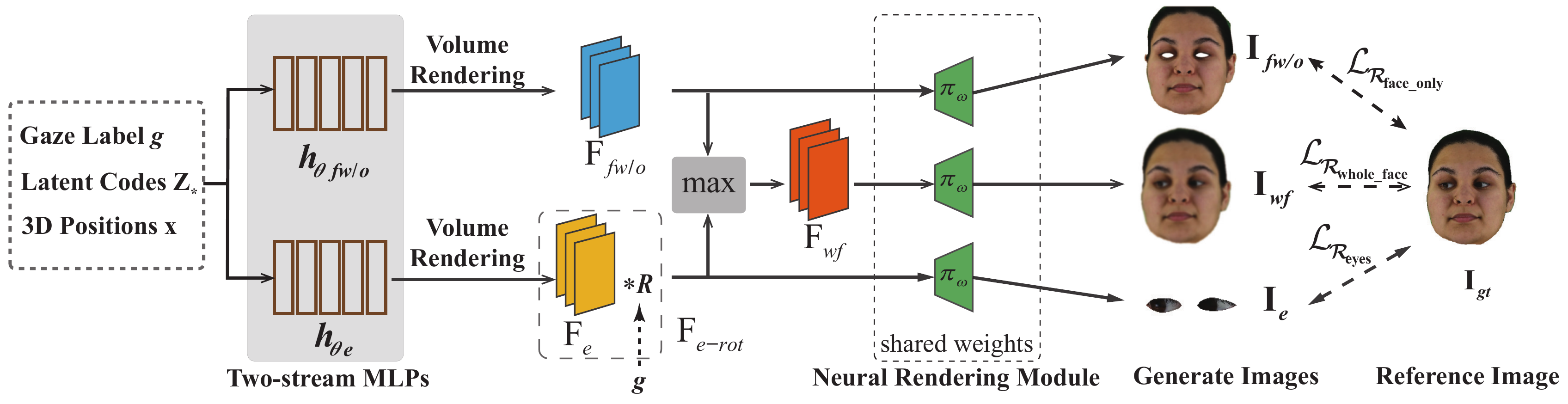}
    \caption{Overview of \methodname pipeline. \methodname trains a two-stream-MLP structure to learn the 3D-aware of the face without eyes feature $F_\text{fw/o}$ and the two eyes feature $F_e$ separately via a NeRF-based model. To model the rigid rotation of two eyeballs, we explicitly multiply $F_e$ with a gaze rotation matrix \textbf{R} to be $F_\text{e-rot}$. The $F_\text{fw/o}$ and $F_\text{e-rot}$ are merged via the max operation to be $F_\text{wf}$. All three features are used to render the face without eyes $\textbf{I}_{\text{fw/o}}$, the eyes $\textbf{I}_{\text{e}}$, and the completed face images $\textbf{I}_{\text{fw}}$.}
    \label{fig:Overview figure of GazeNeRF}
\end{figure*}

\subsection{Recap: NeRF and HeadNeRF}
\label{section3.1}
Neural Radiance Fields, proposed by Mildenhall \etal \cite{nerf}, learns an implicit 3D representation that maps a 3D spatial point \textbf{x} and a view direction \textbf{d} to an RGB color \textbf{c} and a volume density \textbf{$\sigma$}. It parameterizes this continuous implicit function using an MLP as:
\begin{equation}
    h_\theta : (\gamma(\textbf{x}), \gamma(\textbf{d})) \rightarrow (\textbf{c}, \sigma),
\end{equation}
where $\theta$ indicates the network's parameters, and $\gamma$ denotes a positional encoding function~\cite{nerf, vaswani2017attention} transforming $\textbf{x}$ and $\textbf{d}$ into a high-dimensional space.

HeadNeRF \cite{headnerf} is a variant of NeRF for controllable multi-view synthesis and 3D modeling of human faces/heads. Formally, they adjust the MLP as follows:
\begin{equation}
    \begin{aligned}
    h_\theta : (\gamma(\textbf{x}), \textbf{z}_{id}, \textbf{z}_{exp}, \textbf{z}_{alb}, \textbf{z}_{ill}) \rightarrow (\sigma, \textbf{f}).
  \end{aligned}
  \label{eq:head_nerf_mlp}
\end{equation}
Similar to~\cite{niemeyer2021giraffe, gu2021stylenerf}, HeadNeRF replaced the output RGB value with a high-dimensional feature vector $\textbf{f}$. $\textbf{z}_{id}$, $\textbf{z}_{exp}$, $\textbf{z}_{alb}$ and $\textbf{z}_{ill}$ represent the latent codes of the shape, expression, albedo of the face and illumination condition, respectively. The initialization of these latent codes is obtained by fitting the 3D morphable model in~\cite{guo20213d} to the face.

\subsection{GazeNeRF}
\label{section3.2}
We aim to bring 3D awareness to the gaze redirection task by leveraging the high-fidelity image generation and implicit 3D consistency powered by NeRF model.
To this end, we propose \textit{\methodname}, a NeRF-based model with the \mlps and explicit 3D rotation on the eye region.
Motivated by the fact that the face and eyes are two separate physiological entities that can move independently of each other, we propose to use two MLPs instead of one MLP \cite{nerf,headnerf,zhuang2022mofanerf} to separately model the \eyes and \face explicitly, supervised by the segmented image patches.
To introduce a strong 3D prior to the gaze redirection problem, we directly apply the rotation matrix defined by the target gaze direction on the eye stream due to the rigid movement of the eyeballs.
An overview figure of \methodname is presented in Fig.~\ref{fig:Overview figure of GazeNeRF}.

\noindent\textbf{Two-stream MLPs.} 
In contrast to previous gaze redirection methods that mix the \eyes and the \face regions \cite{selflearning,he2019gazeredirection,yu2019improving}, we propose to explicitly disentangle the eyes from the rest of the face with a \mlps to model two separate radiance fields, $h_{\theta_{\text{e}}}$ and $h_{\theta_\text{fw/o}}$ with learnable parameters $\theta_\text{e}$ and $\theta_\text{fw/o}$ for the \eyes and the \face regions respectively. It allows rigid rotation of the two eyeballs along with non-rigid deformation of the periocular areas, such as eyelids and eyebrows. More importantly, it allows for independent control of the transformation or deformation of the two regions.

The two MLPs in \methodname are conditioned on a two-dimensional gaze label consisting of the pitch and yaw angles of the gaze vector in radians, denoted as $g$. In addition, inspired by~\cite{headnerf,zhuang2022mofanerf}, \methodname takes 3DMM parameters as input learnable latent codes to control different factors of the image appearance, such as the shape $\textbf{z}_{sh}$, expression $\textbf{z}_{ex}$, and texture of the face $\textbf{z}_{te}$, and the illumination of the image $\textbf{z}_{il}$.
Both MLPs learn a mapping from 3D locations $\gamma(\textbf{x}) \in \mathbb{R}^{L_\textbf{x}}$ to a generic feature vector $\textbf{f} \in \mathbb{R}^{L_{\textbf{f}}}$ as:
\begin{equation}
    h_{\theta_\text{fw/o}} / h_{\theta_\text{e}} :(\gamma(\textbf{x}), \textbf{z}_{sh}, \textbf{z}_{ex}, \textbf{z}_{te}, \textbf{z}_{il}, g)\rightarrow(\sigma, \textbf{f}).
\end{equation}
With the output from $h_{\theta_\text{fw/o}}$ and $h_{\theta_{\text{e}}}$, we use volume rendering to obtain two low-resolution volume feature maps $F_\text{fw/o}$ and $F_\text{e} \in \mathbb{R}^{64 \times 64 \times 258}$, which are then used to render 2D images.
To ensure each stream generates feature maps corresponding to the correct regions, $F_\text{fw/o}$ and $F_\text{e}$ are later mapped to the segmented \face and \eyes regions, respectively. 

\noindent\textbf{3D awareness.}
Considering that NeRF-based models implicitly learn the 3D volumes of target objects, the feature maps $F_\text{e}$ already incorporate the 3D volume information of the \eyes.
Moreover, previous works incorporate 3D awareness by directly applying the rotation matrix on the 3D volumes \cite{nguyen2020blockgan,nguyen2019hologan}. Such rotation operations also have been shown to work even when rotating in 2D feature space for the gaze redirection task \cite{selflearning,park2019few}. These works, however, apply the rotation to the full face including the eyes ignoring the 3D nature of eyeball rotation and face deformation.
Given the feature maps of the \eyes $F_\text{e}$, we can apply the rotation matrix calculated by the target gaze direction on it to perform the rigid rotation of the eyeball thanks to \mlps disentanglement.
Specifically, we reshape the $F_\text{e} \in \mathbb{R}^{64\times64\times258}$ to $F_\text{e} \in \mathbb{R}^{64\times64\times86\times3}$ and explicitly apply the following transformation to it as $F_\text{e-rot} = \textbf{R}F_\text{e}$, 
where \textbf{R} is a 3D rotation matrix computed from the gaze label $g$ \cite{selflearning,park2019few}. Specifically, we explicitly rotate the feature maps of the \eyes $F_\text{e}$ from the canonical space to $F_\text{e-rot}$ in the target space via a rigid rotation.

\noindent\textbf{Merging features.}
To render the whole face image, we need to combine the feature maps from the two streams, $F_\text{fw/o}$ and $F_\text{e-rot}$.
Similar to how the BlockGAN model combines object features into scene features \cite{nguyen2020blockgan}, we apply the element-wise maximum between $F_\text{fw/o}$ and $F_\text{e-rot}$ to get the merged feature map $F_\text{wf}$. This feature map represents the whole face including both the face and eyes.

\noindent\textbf{Rendering images.} Finally, to render the final 2D images from the feature maps, a neural rendering module~\cite{niemeyer2021giraffe} is adopted. It gradually increases the resolution with learnable upsampling layers. The same strategy is used in~\cite{headnerf,sitzmann2019scene}.
We render the images of the \face region $\textbf{I}_{\text{fw/o}}$ with feature $F_\text{fw/o}$, the \eyes region $\textbf{I}_{\text{e}}$ with feature $F_\text{e-rot}$, and the whole face $\textbf{I}_{wf}$ with feature $F_\text{wf}$. The weights for the rendering module $\pi_\omega$ are shared for all three images.

Given a reference image, we train \methodname and update the learnable parameters including $\theta_\text{e}$ and $\theta_\text{fw/o}$ of \mlps, four latent codes $\textbf{z}_{*}$ and the parameters of the neural rendering module $\pi_{\omega}$
through the minimization of the following objective function:
\begin{equation}
    \mathcal{L}_{Overall}=\lambda_{\mathcal{R}}\mathcal{L}_{\mathcal{R}}+\lambda_{\mathcal{P}}\mathcal{L}_{\mathcal{P}}+\lambda_{\mathcal{F}}\mathcal{L}_{\mathcal{F}}+\lambda_{\mathcal{D}}\mathcal{L}_{\mathcal{D}},
    \label{eq:total_loss}
\end{equation}
where $\mathcal{L}_{\mathcal{R}},\, \mathcal{L}_{\mathcal{P}},\, \mathcal{L}_{\mathcal{F}},\, \mathcal{L}_{\mathcal{D}}$ represent the reconstruction loss, perceptual loss, functional loss, and disentanglement loss, respectively.

\noindent\textbf{Reconstruction Loss.} To generate realistic gaze-redirected images, we apply a reconstruction loss to minimize the pixel-wise distance between a generated image $\textbf{I}_{wf}$ of the whole face and a target image $\textbf{I}_{gt}$, which is formulated as:
\begin{equation}
    \mathcal{L}_{\mathcal{R}_{\text{whole\_face}}}(\textbf{I}_{wf},\textbf{I}_{gt})=\frac{1}{|M_{\textit{wf}}\odot \textbf{I}_{gt}|}||M_{\textit{wf}}\odot(\textbf{I}_{wf}-\textbf{I}_{gt})||_{1},
\label{eq:recon_loss}
\end{equation}
where $M_{\textit{wf}}$ is the whole face mask and $\odot$ stands for a pixel-wise Hadamard product operator. 

To guarantee that the two streams produce $\textbf{I}_{\text{e}}$ and $\textbf{I}_{\text{fw/o}}$ respectively, we apply the similar losses (Eq.~\eqref{eq:recon_loss}) $\mathcal{L}_{\mathcal{R}_{\text{\eyes}}}$ and $\mathcal{L}_{\mathcal{R}_{\text{face\_only}}}$ to the individual images generated by the two streams replacing the whole face mask $M_{\textit{wf}}$ with the \eyes mask $M_{\textit{e}}$ and the \face mask $M_{\textit{f}}$ respectively.
These pixel-wise losses associating masks and images ensure the full disentanglement of the eye and the rest of the face. It further enables us to apply the 3D-aware rotation matrix only to the learned features of the \eyes. It is also helpful to prevent the generation of blurry eyes by applying the reconstruction loss on the \eyes region, since \eyes region is smaller than the \face region.
Hence the final pixel-level reconstruction loss that we use can be written as:
\begin{equation}
    \mathcal{L}_{\mathcal{R}} = \mathcal{L}_{\mathcal{R}_{\text{whole\_face}}} + \mathcal{L}_{\mathcal{R}_{\text{face\_only}}} + \mathcal{L}_{\mathcal{R}_{\text{\eyes}}}.
\end{equation}

\begin{table*}[t]
\centering
\begin{tabularx}{\textwidth}{P{2.0cm} C{1.7cm} C{1.7cm} C{1.7cm} C{1.7cm} C{1.7cm} C{1.7cm} C{1.7cm}}
\toprule
 & Gaze$\downarrow$ & Head Pose$\downarrow$ & SSIM$\uparrow$ & PSNR$\uparrow$ & LPIPS$\downarrow$ & FID$\downarrow$ & Identity Similarity$\uparrow$\\
\midrule
STED & 16.217 & 13.153 & 0.726 & \textbf{17.530} & 0.300 & 115.020 & 24.347 \\
HeadNeRF & 12.117 & 4.275 & 0.720 & 15.298 & 0.294 & \textbf{69.487} & \textbf{46.126} \\
\textbf{\methodname} & \textbf{6.944} & \textbf{3.470} & \textbf{0.733} & 15.453 & \textbf{0.291} & 81.816 & 45.207  \\
\bottomrule
\end{tabularx}
\caption{Comparison of the \methodname to other state-of-the-art methods on the ETH-XGaze dataset in terms of gaze and head redirection errors in degree, redirection image quality (SSIM, PSNR, LPIPS, FID), and identity similarity.}
\label{tab:comare_xgaze}
\end{table*}

\noindent\textbf{Perceptual Loss.} The Perceptual loss~\cite{johnson2016perceptual} is designed to measure perceptual and semantic differences between two images with an image classification network $\phi$, which has been proved effective in previous works \cite{jindal2021cuda,headnerf}. 
We employ a perceptual loss to supervise the generated image $\textbf{I}_{wf}$ to perceptually match with the ground truth image $\textbf{I}_{gt}$,
which is formulated as:
\begin{equation}
    \mathcal{L}_{\mathcal{P}_{\text{whole\_face}}}=\sum_{i}\frac{1}{|\phi_{i}(\textbf{I}_{gt})|}||\phi_{i}(\textbf{I}_{wf})-\phi_{i}(\textbf{I}_{gt})||_{1},
    \label{eq:perc_loss}
\end{equation}
where $i$ denotes the $i$-th layer of VGG16 network~\cite{vgg} pre-trained on ImageNet~\cite{krizhevsky2017imagenet}. Following the same structure of the reconstruction loss, we compute the perceptual losses, $\mathcal{L}_{\mathcal{P}_{\text{\face}}}$ and $\mathcal{L}_{\mathcal{P}_{\text{\eyes}}}$, for the \face and the \eyes images from the two streams, $\textbf{I}_{fw/o }$ and $\textbf{I}_{e}$.
The total perceptual loss is defined as:
\begin{equation}
    \mathcal{L}_{\mathcal{P}}=\mathcal{L}_{\mathcal{P}_{\text{whole\_face}}}+\mathcal{L}_{\mathcal{P}_{\text{\face}}}+\mathcal{L}_{\mathcal{P}_{\text{\eyes}}}.
\end{equation}

\noindent\textbf{Functional Loss.} 
To improve task-specific performance and remove task-relevant inconsistencies between the target image $\textbf{I}_{gt}$ and the reconstructed image $\textbf{I}_{wf}$, we adopt the functional loss from STED \cite{selflearning}. We only include the content-consistency loss formulated as:
\begin{equation}
    \begin{aligned}
    \mathcal{L}_{\mathcal{F}_{\text{content}}} (\textbf{I}_{wf}, \textbf{I}_{gt}) =  \mathcal{E}_{\text{ang}}(\psi^g(\textbf{I}_{wf}), \psi^g(\textbf{I}_{gt}))  ,
    \end{aligned}
\end{equation}
\begin{equation}
    \mathcal{E}_{\text{ang}}(\mathbf{v}, \ \hat{\mathbf{v}})= \arccos  \frac{\mathbf{v} \cdot {\hat{\mathbf{v}}}}{\|\mathbf{v}\| \, \|{\hat{\mathbf{v}}}\|} \, ,
\label{eq:angular_error}
\end{equation}
where $\psi^g(*)$ represents the gaze direction estimated by a gaze estimator network, and $\mathcal{E}_{\text{ang}}(*,*)$ represents the angular error function. Our final functional loss is formulated as follows:
\begin{equation}
    \mathcal{L}_{\mathcal{F}}= \lambda_{\mathcal{F}_{\text{content}}}\mathcal{L}_{\mathcal{F}_{\text{content}}}.
    \label{eq:func_loss}
\end{equation}

\noindent\textbf{Disentanglement Loss.} Inherited from HeadNeRF \cite{headnerf}, to disentangle the effect of the latent codes, we minimize the distance between learned latent codes and the initialization to avoid obvious variations as:
\begin{equation}
    \mathcal{L}_{\mathcal{D}}=\sum\frac{w_{*}}{|\textbf{z}_{*}^{0}|}||\textbf{z}_{*}-\textbf{z}_{*}^{0}||^{2},
    \label{eq:dis_loss}
\end{equation}
where $\textbf{z}_{*}^{0}$ denotes the initial values of the four latent codes obtained from 3DMM parameters, and $w_{*}$ represents the loss weight.
\section{Experiments}
To demonstrate the effectiveness of \methodname, we first train \methodname on the ETH-XGaze dataset~\cite{xgaze} and compare it to the current state-of-the-art gaze redirection and face generation methods with multiple evaluation metrics. We then conduct cross-dataset evaluations with key evaluation metrics to show the generalization of \methodname. We further analyze the contribution of the various individual components of \methodname to the performance with an ablation study.

\begin{figure*}[htbp]
    \centering
    \includegraphics[width=0.97\textwidth]{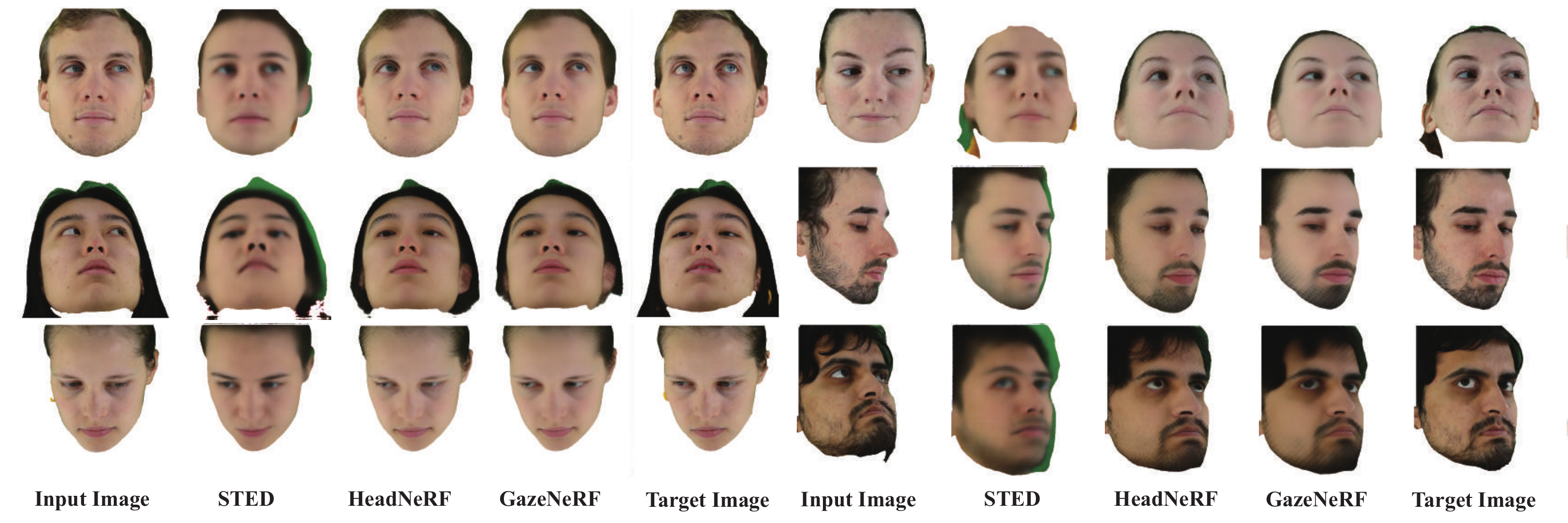}
    \caption{Visualization of generated images from ETH-XGaze with our \methodname, STED and HeadNeRF. All faces are applied with face masks to remove the background. Our \methodname can generate photo-realistic face images with different gaze directions and head poses. STED suffers from losing identity information, and HeadNeRF cannot generate fine-grained eyes (the second row).}
    \label{fig:visualization}
\end{figure*}

\subsection{Datasets}
\noindent\textbf{ETH-XGaze}~\cite{xgaze} is a large-scale gaze dataset of high-resolution images with extreme head pose and gaze variation, which was acquired under a multi-view camera system with different illumination conditions. 
There are 756K frames of 80 subjects in the training set. Each frame is composed of 18 different camera view images. The person-specific test set contains 15 subjects
with two hundred images from each subject provided with ground truth gaze labels. 
\newline
\textbf{MPIIFaceGaze}~\cite{overface} is an additional dataset for appearance-based gaze estimation based on MPIIGaze dataset~\cite{deepappearence}. MPIIFaceGaze provides 3000 face images with two-dimensional gaze labels for every 15 subjects.
\newline
\textbf{ColumbiaGaze}~\cite{CAVE_0324} consists of 5880 high-resolution images taken from 56 subjects. For each subject, the images were acquired with the same five fixed head poses and 21 fixed gaze directions per head pose.
\newline
\textbf{GazeCapture}~\cite{krafka2016eye} is a large-scale dataset were taken with different poses and under different illumination conditions via crowd-sourcing. During the cross-dataset evaluation, we use its test set only, which contains 150 subjects.
\subsection{Implementation details}
We employ Adam~\cite{adam} as our optimizer with $1e^{-4}$ as the learning rate.
We use a VGG-based \cite{Simonyan15} network pre-trained on ImageNet and fine-tune it on the ETH-XGaze training set for the functional loss ${L}_{\mathcal{F}}$ as the pre-trained gaze estimator.
We train another ResNet50 backbone as in~\cite{he2016deep} on the ETH-XGaze training set that outputs gaze and head pose for evaluation purposes.
Finally, we empirically set the total loss coefficients in equation \eqref{eq:total_loss} to  $\lambda_{\mathcal{R}} = \lambda_{\mathcal{P}} = \lambda_{\mathcal{F}} = \lambda_{\mathcal{D}} = 1$, and the disentanglement weights in equation \eqref{eq:dis_loss} to $w_{\text{sh}} = w_{\text{te}} = w_{\text{il}} = \num{1e-3}$ and $w_{\text{ex}} = 1.0$. While $\lambda_{\mathcal{F}_{\text{content}}}$ in equation \eqref{eq:func_loss} is set to $\num{1e-3}$ and is increased by $\num{1e-3}$ after each epoch.

\begin{table*}[t]
\centering
\begin{tabularx}{\textwidth}{P{1.8cm} | C{0.8cm} C{0.8cm} C{0.8cm} C{1.0cm} | C{0.8cm} C{0.8cm} C{0.8cm} C{1.0cm} | C{0.8cm} C{0.8cm} C{0.8cm} C{1.0cm}}
\toprule
 & \multicolumn{4}{c}{ColumbiaGaze} & \multicolumn{4}{c}{MPIIFaceGaze} & \multicolumn{4}{c}{GazeCapture} \\
 & Gaze$\downarrow$ & Head$\downarrow$ & LPIPS$\downarrow$ & ID$\uparrow$ & Gaze$\downarrow$ & Head$\downarrow$ & LPIPS$\downarrow$ & ID$\uparrow$ & Gaze$\downarrow$ & Head$\downarrow$ & LPIPS$\downarrow$ & ID$\uparrow$ \\
\midrule

STED & 17.887 & 14.693 & 0.413 & 6.384 & 14.796 & 11.893 & 0.288 & 10.677 & 15.478 & 16.533 & 0.271 & 6.807 \\
HeadNeRF & 15.250 & 6.255 & \textbf{0.349} & \textbf{23.579} & \textbf{14.320} & 9.372 & 0.288 & \textbf{31.877} & 12.955 & 10.366 & \textbf{0.232} & \textbf{20.981} \\
\midrule
\textbf{\methodname} & \textbf{9.464} & \textbf{3.811} & 0.352 & 23.157 & 14.933 & \textbf{7.118} & \textbf{0.272} & 30.981 & \textbf{10.463} & \textbf{9.064} & \textbf{0.232} & 19.025 \\
\bottomrule
\end{tabularx}
\caption{Comparison of \methodname to other state-of-the-art methods on ColumbiaGaze, MPIIFaceGaze, and GazeCapture datasets in terms of gaze and head redirection errors in degree, LPIPS, and Identity similarity (ID).}
\label{tab:comare_otherdataset}
\end{table*}

\subsection{Experimental setup}
\noindent\textbf{Pre-processing procedure.} 
We apply the data normalization method~\cite{zhang18_etra,sugano2014learning} and resize the face images into 512$\times$512 pixels. To guarantee that our \mlps architecture learns to render the \face and \eyes regions separately, we utilize face parsing models~\cite{faceparsingmodel} to generate masks for them. We also adopt the 3D face parametric model from~\cite{guo20213d} to generate the four latent codes as input into our model. We convert the provided gaze labels from all datasets into pitch-yaw angle labels in the head coordinate system for consistency across subjects and datasets. See the details in the supplementary.
\newline
\textbf{Evaluation metrics.} We evaluate all models with various metrics, which can be divided into three different categories: redirection error, redirection image quality, and identity similarity. 
Similar to STED, redirection error is composed of gaze and head pose angular errors estimated by the ResNet50~\cite{he2016deep}-based estimator. These errors are measured with the estimated gaze directions between the redirected images and the corresponding ground truth images. 
To measure the quality of reconstructed images, we adopt four different metrics, including Structure Similarity Index(SSIM), Peak Signal-to-Noise Ratio(PSNR), Learned Perceptual Image Patch Similarity(LPIPS), and Fr\'echet Inception Distance(FID). Identity similarity is measured based on the face recognition model from FaceX-Zoo~\cite{wang2021facex}. It measures the differences in identity between the redirected images and ground truth images.

\subsection{Comparison to state of the art}
To show the superiority of \methodname, we compare \methodname with several previous works in two different experiments: within-dataset evaluation in Tab.~\ref{tab:comare_xgaze} and cross-dataset evaluation in Tab.~\ref{tab:comare_otherdataset}. In both experiments, all models are trained with 14.4K images from 10 frames per subject, 18 camera view images per frame, and 80 subjects on the ETH-XGaze training set.\\
\newline
\textbf{Methods.} 
We compare against the existing state-of-the-art gaze redirection model STED~\cite{selflearning}, and other variants of the NeRF-based HeadNeRF~\cite{headnerf} models that could also be modified to redirect gaze. 
STED is the current state-of-the-art gaze redirection method applied to full-face images. It performs better than previous works from He \cite{he2019gazeredirection}, DeepWarp \cite{deepwarp} and StarGAN \cite{choi2018stargan}. HeadNeRF is a NeRF-based method that generates high-fidelity face images with 3DMM latent codes controlling different factors of faces. We adapted STED to our setting by increasing the input and target images dimension from 128$\times$128 to 512$\times$512 pixels. 
To adapt the NeRF-based model for gaze redirection, we concatenated the two-dimensional gaze labels to the original inputs of HeadNeRF directly and conditioned the MLP to learn gaze-related information.
\subsubsection{Within-dataset evaluation}
Since \methodname and other methods require the gaze label as input, we evaluate their performance on the person-specific test set of ETH-XGaze. There are 15 subjects in the person-specific test set, where 200 images per subject have been annotated with head pose and gaze labels. We randomly pair these 200 labeled images per subject as input and target images. The same pairs of images are evaluated for all models. 

Tab.~\ref{tab:comare_xgaze} shows the evaluation results of \methodname and other methods. From the table, we can see that \methodname achieves better results than STED and HeadNeRF for most of the error metrics. Especially, \methodname achieves the best performance on the gaze and head redirection as the core criteria of a gaze redirection method.
Compared to the HeadNeRF conditioned on the gaze label with single MLP, \methodname applies an explicit rotation to the learned feature maps of the two eyes, which provides better control of gaze direction with smaller gaze error.
Although also explicitly applies rotation to the feature maps, STED performs worse than \methodname in terms of gaze and head pose errors. It is because STED utilizes 2D generative model that lacks 3D awareness in its feature maps, and it does not separate the eyes from the face. \methodname achieves similar results as HeadNeRF in terms of the image quality error metrics SSIM, PSNR and LPIPS. This shows that the increase in gaze redirection accuracy does not come at the cost of image fidelity. 

We show a qualitative comparison of the various methods in Fig.~\ref{fig:visualization}. It clearly shows that \methodname generates photo-realistic face images for variant gaze directions and head poses. STED suffers from the loss of personal identity information in the generated face images, which is quantitatively verified as the `identity similarity' in Tab.~\ref{tab:comare_xgaze}. Moreover, STED has difficulty in dealing with extreme head poses (the second row left and the first row right), where the generated faces shift from the target poses. 
As for HeadNeRF, the feature maps from the single MLP conditioned on gaze labels as inputs alone are not strong enough to control the appearance of eyes with various gaze directions (the last row). Even though most of the results of HeadNeRF can preserve the face identity, the rest of them fail to generate fine-grained eyes (the second row). Compared with these two state-of-the-art methods, \methodname generates better face images with fine-grained eyes, even with extreme head poses (the middle two rows from right).

\subsubsection{Cross-dataset evaluation}
To evaluate the generalization of \methodname, we conduct a cross-dataset evaluation. For the cross-dataset evaluation, we train the same methods as for the within-dataset evaluation and test on three other datasets, namely ColumbiaGaze, MPIIFaceGaze, and the test set of GazeCapture. Similar to the within-dataset evaluation, we randomly pair the images per subject and fix the pairs for all models. We adopt gaze and head pose angular errors, LPIPS and identity similarity as the evaluation metrics.

The results from Tab.~\ref{tab:comare_otherdataset} show that \methodname achieves the best performance on the three datasets for most evaluation metrics. 
As the same as the within-dataset evaluation, \methodname significantly outperforms the other two methods in terms of gaze angular and head pose errors with big margins only except the gaze error in the MPIIFaceGaze dataset.
Compared to the HeadNeRF, \methodname achieves better performance in terms of the head pose redirection, although the head rotation operation is the same for both methods.
It could be because the \mlps architecture adds additional ability control for the \face region by separating the face and eye.
All three models have similar performance in image quality as in Tab.~\ref{tab:comare_xgaze}. STED still suffers from the loss of personal identity.

\subsection{Ablation study}
\begin{table*}[t]
\centering
\begin{tabularx}{\textwidth}{P{4.5cm} C{1.3cm} C{1.3cm} C{1.3cm} C{1.3cm} C{1.3cm} C{1.3cm} C{1.3cm} }
\toprule
 & Gaze$\downarrow$ & Head Pose$\downarrow$ & SSIM$\uparrow$ & PSNR$\uparrow$ & LPIPS$\downarrow$ & FID$\downarrow$ & Identity Similarity$\uparrow$\\
\midrule
vanilla-GazeNeRF & 11.427 & 4.581 & 0.722 & 15.254 & 0.291 & 71.971 & 47.751\\
vanilla-GazeNeRF+rotation & 9.279 & 4.458 & 0.724 & 15.273 & 0.296 & 75.112 & 47.642 \\
Two-stream & 8.609 & 3.527 & 0.731 & 15.431 & \textbf{0.286} & \textbf{69.339} & \textbf{48.649} \\
Two-stream+rotation & 8.437 & 3.563 & 0.730 & 15.368 & 0.296 & 79.289 & 48.284 \\
vanilla-GazeNeRF+${L}_{\mathcal{F}}$ & 7.777 & 4.127 & 0.729 & 15.404 & 0.306 & 92.201 & 38.395 \\
\textbf{\methodname (Two-stream + rotation + ${L}_{\mathcal{F}}$)} & \textbf{6.944} & \textbf{3.470} & \textbf{0.733} & \textbf{15.453} & 0.291 & 81.816 & 45.207 \\
\bottomrule
\end{tabularx}
\caption{Comparison of \methodname to its other variations on the ETH-XGaze dataset in terms of gaze and head redirection errors in degree, redirection image quality (SSIM, PSNR, LPIPS and FID), and identity similarity.}
\label{tab:comare_ablation}
\end{table*}

We analyze \methodname through a number of ablation experiments in Tab.~\ref{tab:comare_ablation}. We show the strength of \methodname by comparing it with alternative design choices as listed in the following.

\noindent\textbf{\textit{Vanilla-GazeNeRF}.} We utilize a single MLP and concatenate the two-dimensional gaze label with its other inputs instead of adopting our proposed \mlps and adding a 3D-aware rotation matrix. Different from the HeadNeRF from Tab.~\ref{tab:comare_xgaze}, we adopt $\mathcal{L}1$ reconstruction loss used in \methodname instead of $\mathcal{L}2$ photometric loss used in the original HeadNeRF~\cite{headnerf} for a fair comparison. During training, all training objectives except functional loss are used.
\newline
\textbf{3D awareness.} To verify the individual contributions of the various components of \methodname, namely its \mlps architecture and its 3D-aware rotation, we train three more models, \textit{vanilla GazeNeRF+rotation}, \textit{Two-stream} and \textit{Two-stream+rotation} for comparison. 
Similar to the previous work, STED and FAZE, both of which apply the rotation matrix to the intermediate feature maps of the whole face, \textit{vanilla-GazeNeRF + rotation} applies the gaze rotation matrix to the feature maps of the single MLP. 
\textit{Two-stream} has a two-stream MLP structure for generating the face without two eyes and two-eye regions separately. Both streams only take the gaze label as input without applying a rotation matrix to the feature maps of two eyes. 
\textit{Two-stream + rotation} multiplies the 3D-aware feature maps from the two eyes stream with the rotation matrix. Similar to \textit{vanilla-GazeNeRF}, the functional loss is not used for optimizing \textit{Two-stream} and \textit{Two-stream + rotation}.
\newline
\textbf{Functional loss.} \textit{Vanilla-GazeNeRF+${L}_{\mathcal{F}}$} is trained to verify the power of the functional loss for the gaze redirection task. Compared to \textit{vanilla-GazeNeRF}, the functional loss is used along with the other losses.

From the results shown in Tab.~\ref{tab:comare_ablation}, we can see that the baseline model \textit{vanilla-GazeNeRF} performs the worst in terms of gaze error. Comparing \textit{vanilla-GazeNeRF+rotation} with the \textit{vanilla-GazeNeRF}, both the gaze and the head pose angular errors drop. The performance of two angular errors profits from the rotation matrix is applied to the feature maps incorporating the information of the whole face. Moreover, the smaller gaze and head pose angular errors of \textit{Two-stream} are due to the two-stream-MLP structure that separates the whole face into the \face and \eyes parts. We can also see that applying rotation matrix to the \eyes stream on the basis of \textit{Two-stream} benefits both angular errors of \textit{Two-stream+rotation}. In addition, adding the functional loss ${L}_{\mathcal{F}}$ as shown with \textit{Vanilla-GazeNeRF+${L}_{\mathcal{F}}$} improves the gaze error greatly since it uses an additional gaze estimator to minimize the gaze-relevant inconsistency between the generated and ground-truth images.

Among all ablations, \methodname achieves the best performance in terms of gaze and head pose angular errors by taking advantage of the combination of two-stream-MLP structure, applying a rotation matrix to the \eyes stream, and using the function loss ${L}_{\mathcal{F}}$. As for the image quality, \methodname achieves the best performance regarding SSIM and PSNR score and is comparable to best performances with slight differences for image quality and identify similarities metrics. Again, we emphasize that our goal is not to improve the overall image quality but rather to improve gaze redirection accuracy. 
\section{Conclusion and Discussion}
We propose \methodname, the first method that introduces 3D awareness to the gaze redirection task. 
By considering the 3D nature of the gaze redirection task itself, \methodname consists of a \mlps and explicit rotation on the disentangled eye volumes feature.
The 3D-aware design endows the advantage of \methodname for the gaze redirection task, which has been proven by the leading performance on multiple datasets and ablation studies.
We believe \methodname has great potential for downstream applications with the benefits of 3D awareness.
Notwithstanding the above advantages, \methodname shares the same limitation of the group of NeRF models that it takes a long time to train. We leave reducing the burden of training time as our future work.
\section*{Acknowledgement}
This work was supported by Institute of Information \& communications Technology Planning \& Evaluation (IITP) grant funded by the Korea government(MSIT) (No.2022-0-00608, Artificial intelligence research about multi-modal interactions for empathetic conversations with humans)

\setcounter{equation}{0}
\setcounter{figure}{0}
\setcounter{table}{0}
\setcounter{page}{1}
\setcounter{section}{0}

\clearpage
\title{GazeNeRF: 3D-Aware Gaze Redirection with Neural Radiance Fields \\
\large{Supplementary Material}}

\author{}
\date{}

\maketitle
\pagestyle{empty}
\thispagestyle{empty}

\section{Overview}
In this supplementary material, we first provide more details of the data pre-processing and training procedure. We also show another ablation study results on loss components and a comparison between \methodname trained from scratch and the pre-trained HeadNeRF~\cite{headnerf} model. We then show additional qualitative results on different datasets. Furthermore, we show the results of the few-shot personal calibration experiments. We encourage the readers to also watch the supplementary video that contains more animated results of the proposed method.

\section{Details of data pre-processing and training precedure}
The original resolution of images from ETH-XGaze~\cite{xgaze} is 6K$\times$4K, and the resolutions of the images from other datasets are different from each other. To unify them, we pre-process the images with the data normalization method in~\cite{zhang18_etra}, where the rotation and translation between the camera and face coordinate systems are standardized. We fix the normalized distance between the camera and the center of the face to 680mm. To centralize the faces in the normalized images, we use different values for the focal lengths for the normalized camera projection matrices, which are 1600, 1400, 1600 and 1200 for ETH-XGaze~\cite{xgaze}, MPIIFaceGaze~\cite{zhang2017mpiigaze}, ColumbiaGaze~\cite{CAVE_0324} and GazeCapture~\cite{krafka2016eye}, respectively.

To obtain the 3DMM parameters and the masks of the \eyes and the \face regions, we use the face parsing model in~\cite{faceparsingmodel} to segment the whole face. For some images, we also use the face parsing model in~\cite{CelebAMask-HQ} and facial landmarks~\cite{bulat2017far} to determine the eye masks only when the face paring model~\cite{faceparsingmodel} returns empty results for the eyes.

\methodname is trained with a single NVIDIA A40 GPU for one week. During inference, we fine-tune GazeNeRF and update four learable latent codes using a single image. Fine-tuning takes around one minute, and generating new image in one second.

\section{Ablations on loss components}
In this section, we show another ablation study on the contributions of different loss components. We train another three baseline \methodname with different loss components. Here the baseline \methodname represents the structure of \methodname, which is \textit{Two-stream+rotation}. We take the reconstruction loss as the base and verify the power of different loss components in an additive way. The results are listed in Tab.~\ref{tab:loss_ablation} and evaluated on ETH-XGaze dataset.

The results show that only using reconstruction loss achieves the worst performance regarding all evaluation metrics. Adding the perceptual loss boosts the performance in all metrics, especially gaze and head pose angular errors. Moreover, adding the disentanglement loss achieves the best performance in the most of evaluation metrics. Utilizing the functional loss helps to drop the gaze angular error of \methodname at the cost of image quality (\eg FID) and person identity.

\section{Comparison between \methodname trained from scratch and the pre-trained model}

Tab.~\ref{tab:comare_pretrained} shows the evaluation results between \methodname trained from scratch and the pre-trained HeadNeRF model~\cite{headnerf}. We can find that training with the pre-trained model helps improve the head pose error at the cost of the gaze angular error. Regarding the image quality and identity similarity, both models conduct the similar performance.

\section{Personal calibration for gaze estimation}

\begin{figure}[t]
    \centering  \includegraphics[width=0.48\textwidth]{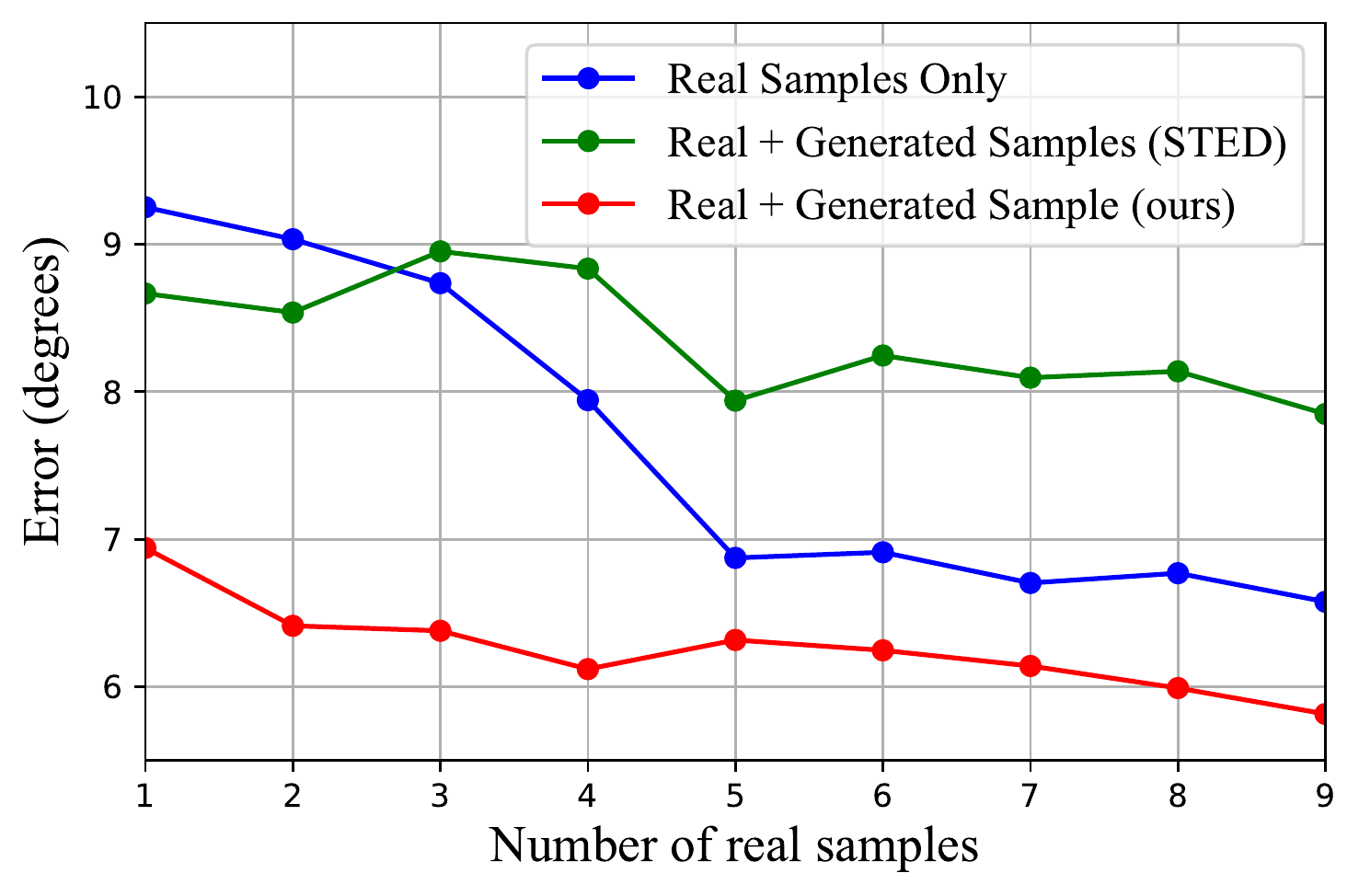}
    \caption{Downstream personal gaze estimation task in a few-shot setting. The x-axis is the number of real samples used, and the y-axis is the gaze estimation error in degree. The results are calculated by averaging the angular error between the 15 subjects of the ETH-XGaze person-specific set. We show the result of only using real samples (blue), using real plus generated samples from STED (green), and using real plus generated samples from our GazeNeRF (red) to fine-tune the pre-trained gaze estimator.}
    \label{fig:personspecific}
\end{figure}

\begin{table*}[ht]
\centering
\begin{tabularx}{\textwidth}{P{4.5cm} C{1.3cm} C{1.3cm} C{1.3cm} C{1.3cm} C{1.3cm} C{1.3cm} C{1.3cm} }
\toprule
 & Gaze$\downarrow$ & Head Pose$\downarrow$ & SSIM$\uparrow$ & PSNR$\uparrow$ & LPIPS$\downarrow$ & FID$\downarrow$ & Identity Similarity$\uparrow$\\
\midrule
Baseline \methodname+$\mathcal{L}_{\mathcal{R}}$ & 28.122 & 21.489 & 0.683 & 14.150 & 0.406 & 146.943 & 20.499 \\
Baseline \methodname+$\mathcal{L}_{\mathcal{R}}$+$\mathcal{L}_{\mathcal{P}}$ & 8.861 & 3.456 & 0.729 & 15.370 & 0.290 & 72.133 & \textbf{49.265} \\
Baseline \methodname+$\mathcal{L}_{\mathcal{R}}$+$\mathcal{L}_{\mathcal{P}}$+$\mathcal{L}_{\mathcal{D}}$ & 8.460 & \textbf{3.386} & 0.729 & \textbf{15.461} & \textbf{0.288} & \textbf{72.044} & 48.705 \\
\midrule
\methodname (Baseline \methodname+$\mathcal{L}_{\mathcal{R}}$+$\mathcal{L}_{\mathcal{P}}$+$\mathcal{L}_{\mathcal{D}}$+${L}_{\mathcal{F}}$) & \textbf{6.944} & 3.470 & \textbf{0.733} & 15.453 & 0.291 & 81.816 & 45.207 \\
\bottomrule
\end{tabularx}
\caption{Ablation study on different loss components.}
\label{tab:loss_ablation}
\end{table*}

\begin{table*}[t]
\centering
\begin{tabularx}{\textwidth}{P{4.5cm} C{1.3cm} C{1.3cm} C{1.3cm} C{1.3cm} C{1.3cm} C{1.3cm} C{1.3cm} }
\toprule
 & Gaze$\downarrow$ & Head Pose$\downarrow$ & SSIM$\uparrow$ & PSNR$\uparrow$ & LPIPS$\downarrow$ & FID$\downarrow$ & Identity Similarity$\uparrow$\\
\midrule
\textbf{\methodname (Two-stream + rotation + ${L}_{\mathcal{F}}$)} & \textbf{6.944} & 3.470 & \textbf{0.733} & \textbf{15.453} & 0.291 & 81.816 & \textbf{45.207} \\
\textbf{\methodname} + pre-trained HeadNeRF & 7.134 & \textbf{3.080} & 0.732 & 14.761 & \textbf{0.285} & \textbf{76.293} & 43.443 \\
\bottomrule
\end{tabularx}
\caption{Comparison of \methodname trained from scratch to pre-trained HeadNeRF model.}
\label{tab:comare_pretrained}
\end{table*}

In this section, we demonstrate how \methodname is beneficial for the downstream task of person-specific gaze estimation in a few-shot setting. 
Specifically, given a few calibration samples from person-specific test sets, we augment these real samples with gaze redirected samples generated by \methodname. 
We then fine-tune the gaze estimator pre-trained on ETH-XGaze's training set with these augmented samples and compare the performance with the baseline model that is fine-tuned only with real samples. 
To eliminate the influence of the number of samples, the size of augmented samples is always 200 (real + generated samples). We change the number of real samples used for the fine-tuning during the evaluation.

The result is shown in Fig.~\ref{fig:personspecific}, where the x-axis is the number of real samples used and the y-axis is the gaze estimation error in degree on the ETH-XGaze person-specific test set. We test up to nine real samples for the few-shot setting.
Observe from the figure that fine-tuning the pre-trained gaze estimator with both real and generated samples from \methodname brings a significant improvement in gaze error versus only fine-tuning with real samples. This trend is more evident when fewer real samples are available. It indicates that the generated sample from \methodname is of high fidelity in terms of its gaze angle, such that it can be helpful to improve the downstream gaze estimation accuracy. In Fig. \ref{fig:personspecific} we also compare the result of few-shot personal calibration when the generated samples come from STED~\cite{selflearning}. ST-ED performs the worst in this case. It shows that the 2D generative model is less helpful for the downstream gaze estimation task, which is due to the lack of consideration of the 3D nature of the gaze redirection task.

\section{Additional qualitative results}

\begin{figure*}
    \centering
    \includegraphics[width=0.89\textwidth]{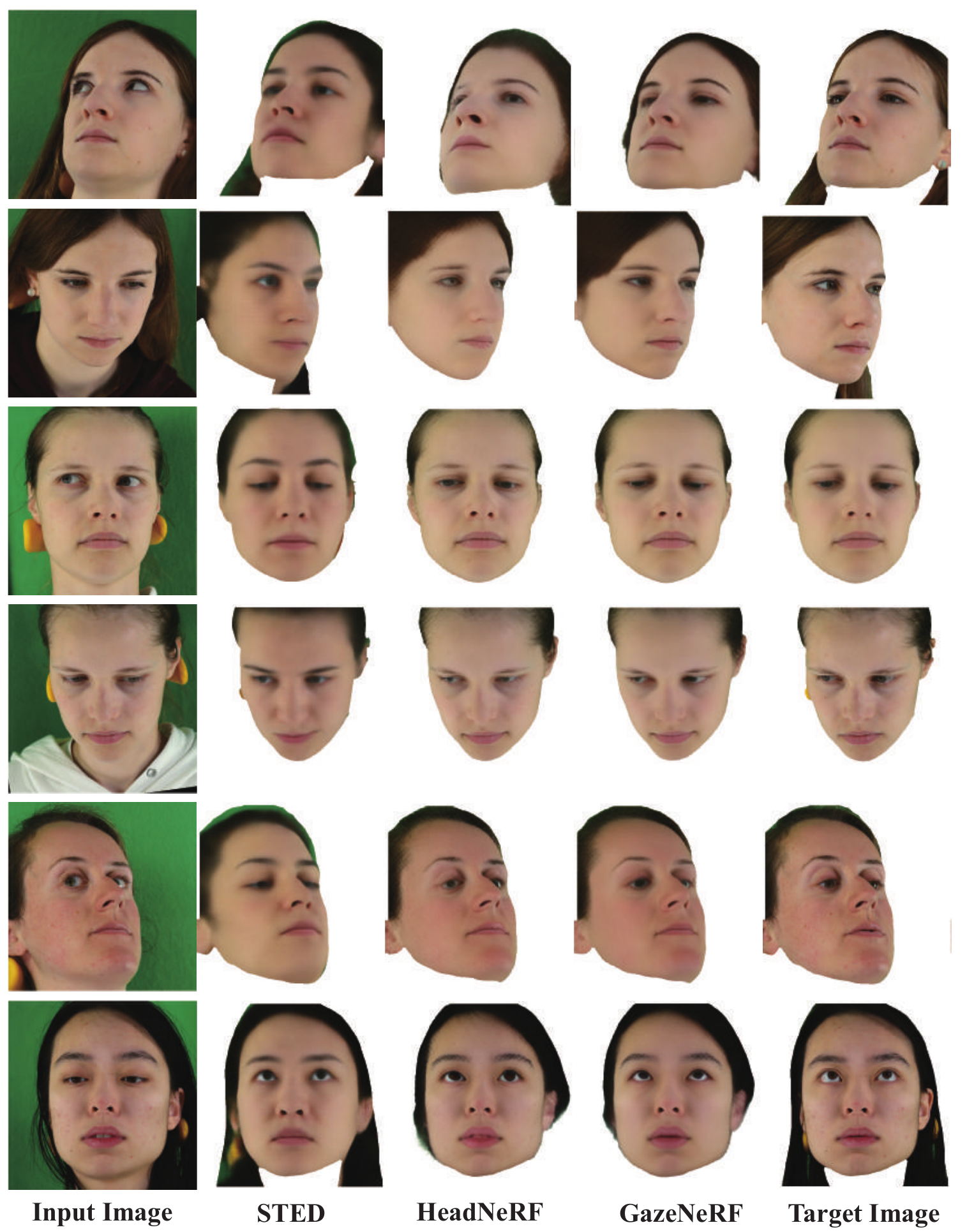}
    \caption{Additional visualization of generated images from ETH-XGaze with our \methodname, STED and HeadNeRF. All faces are applied with face masks to remove the background. Our \methodname can generate photo-realistic face images with different gaze directions and head poses. STED suffers from losing identity information, and HeadNeRF cannot generate fine-grained eyes.}
    \label{fig:visualization_appendix}
\end{figure*}

In Fig.~\ref{fig:visualization_appendix} we show additional qualitative results of \methodname and the SOTA baselines, evaluated on the person-specific test set of the ETH-XGaze dataset.

{\small
\bibliographystyle{ieee_fullname}
\bibliography{08_references}

\begin{thebibliography}{10}\itemsep=-1pt

\bibitem{bulat2017far}
Adrian Bulat and Georgios Tzimiropoulos.
\newblock How far are we from solving the 2d \& 3d face alignment problem? (and
  a dataset of 230,000 3d facial landmarks).
\newblock In {\em International Conference on Computer Vision}, 2017.

\bibitem{choi2018stargan}
Yunjey Choi, Minje Choi, Munyoung Kim, Jung-Woo Ha, Sunghun Kim, and Jaegul
  Choo.
\newblock Stargan: Unified generative adversarial networks for multi-domain
  image-to-image translation.
\newblock In {\em Proceedings of the IEEE conference on computer vision and
  pattern recognition}, pages 8789--8797, 2018.

\bibitem{gafni2021dynamic}
Guy Gafni, Justus Thies, Michael Zollhofer, and Matthias Nie{\ss}ner.
\newblock Dynamic neural radiance fields for monocular 4d facial avatar
  reconstruction.
\newblock In {\em Proceedings of the IEEE/CVF Conference on Computer Vision and
  Pattern Recognition}, pages 8649--8658, 2021.

\bibitem{deepwarp}
Yaroslav Ganin, Daniil Kononenko, Diana Sungatullina, and Victor Lempitsky.
\newblock Deepwarp: Photorealistic image resynthesis for gaze manipulation.
\newblock In {\em European conference on computer vision}, pages 311--326.
  Springer, 2016.

\bibitem{gu2021stylenerf}
Jiatao Gu, Lingjie Liu, Peng Wang, and Christian Theobalt.
\newblock Stylenerf: A style-based 3d aware generator for high-resolution image
  synthesis.
\newblock In {\em International Conference on Learning Representations}, 2022.

\bibitem{guo20213d}
Yudong Guo, Lin Cai, and Juyong Zhang.
\newblock 3d face from x: Learning face shape from diverse sources.
\newblock {\em IEEE Transactions on Image Processing}, 30:3815--3827, 2021.

\bibitem{he2016deep}
Kaiming He, Xiangyu Zhang, Shaoqing Ren, and Jian Sun.
\newblock Deep residual learning for image recognition.
\newblock In {\em Proceedings of the IEEE conference on computer vision and
  pattern recognition}, pages 770--778, 2016.

\bibitem{he2019gazeredirection}
Zhe He, Adrian Spurr, Xucong Zhang, and Otmar Hilliges.
\newblock Photo-realistic monocular gaze redirection using generative
  adversarial networks.
\newblock In {\em {IEEE} International Conference on Computer Vision ({ICCV})}.
  {IEEE}, 2019.

\bibitem{headnerf}
Yang Hong, Bo Peng, Haiyao Xiao, Ligang Liu, and Juyong Zhang.
\newblock Headnerf: A real-time nerf-based parametric head model.
\newblock In {\em Proceedings of the IEEE/CVF Conference on Computer Vision and
  Pattern Recognition}, pages 20374--20384, 2022.

\bibitem{jang2021codenerf}
Wonbong Jang and Lourdes Agapito.
\newblock Codenerf: Disentangled neural radiance fields for object categories.
\newblock In {\em Proceedings of the IEEE/CVF International Conference on
  Computer Vision}, pages 12949--12958, 2021.

\bibitem{jindal2021cuda}
Swati Jindal and Xin~Eric Wang.
\newblock Cuda-ghr: Controllable unsupervised domain adaptation for gaze and
  head redirection.
\newblock {\em arXiv preprint arXiv:2106.10852}, 2021.

\bibitem{johnson2016perceptual}
Justin Johnson, Alexandre Alahi, and Li Fei-Fei.
\newblock Perceptual losses for real-time style transfer and super-resolution.
\newblock In {\em European conference on computer vision}, pages 694--711.
  Springer, 2016.

\bibitem{adam}
Diederik~P. Kingma and Jimmy Ba.
\newblock Adam: {A} method for stochastic optimization.
\newblock In {\em 3rd International Conference on Learning Representations,
  {ICLR}}, 2015.

\bibitem{krafka2016eye}
Kyle Krafka, Aditya Khosla, Petr Kellnhofer, Harini Kannan, Suchendra
  Bhandarkar, Wojciech Matusik, and Antonio Torralba.
\newblock Eye tracking for everyone.
\newblock In {\em Proceedings of the IEEE conference on computer vision and
  pattern recognition}, pages 2176--2184, 2016.

\bibitem{krizhevsky2017imagenet}
Alex Krizhevsky, Ilya Sutskever, and Geoffrey~E Hinton.
\newblock Imagenet classification with deep convolutional neural networks.
\newblock {\em Communications of the ACM}, 60(6):84--90, 2017.

\bibitem{lazova2022control}
Verica Lazova, Vladimir Guzov, Kyle Olszewski, Sergey Tulyakov, and Gerard
  Pons-Moll.
\newblock Control-nerf: Editable feature volumes for scene rendering and
  manipulation.
\newblock {\em arXiv preprint arXiv:2204.10850}, 2022.

\bibitem{CelebAMask-HQ}
Cheng-Han Lee, Ziwei Liu, Lingyun Wu, and Ping Luo.
\newblock Maskgan: Towards diverse and interactive facial image manipulation.
\newblock In {\em IEEE Conference on Computer Vision and Pattern Recognition
  (CVPR)}, 2020.

\bibitem{li2022eyenerf}
Gengyan Li, Abhimitra Meka, Franziska Mueller, Marcel~C Buehler, Otmar
  Hilliges, and Thabo Beeler.
\newblock Eyenerf: a hybrid representation for photorealistic synthesis,
  animation and relighting of human eyes.
\newblock {\em ACM Transactions on Graphics (TOG)}, 41(4):1--16, 2022.

\bibitem{nerf}
Ben Mildenhall, Pratul~P. Srinivasan, Matthew Tancik, Jonathan~T. Barron, Ravi
  Ramamoorthi, and Ren Ng.
\newblock Nerf: Representing scenes as neural radiance fields for view
  synthesis, 2020.

\bibitem{nguyen2019hologan}
Thu Nguyen-Phuoc, Chuan Li, Lucas Theis, Christian Richardt, and Yong-Liang
  Yang.
\newblock Hologan: Unsupervised learning of 3d representations from natural
  images.
\newblock In {\em Proceedings of the IEEE/CVF International Conference on
  Computer Vision}, pages 7588--7597, 2019.

\bibitem{nguyen2020blockgan}
Thu~H Nguyen-Phuoc, Christian Richardt, Long Mai, Yongliang Yang, and Niloy
  Mitra.
\newblock Blockgan: Learning 3d object-aware scene representations from
  unlabelled images.
\newblock {\em Advances in Neural Information Processing Systems},
  33:6767--6778, 2020.

\bibitem{niemeyer2021giraffe}
Michael Niemeyer and Andreas Geiger.
\newblock Giraffe: Representing scenes as compositional generative neural
  feature fields.
\newblock In {\em Proceedings of the IEEE/CVF Conference on Computer Vision and
  Pattern Recognition}, pages 11453--11464, 2021.

\bibitem{park2021nerfies}
Keunhong Park, Utkarsh Sinha, Jonathan~T. Barron, Sofien Bouaziz, Dan~B
  Goldman, Steven~M. Seitz, and Ricardo Martin-Brualla.
\newblock Nerfies: Deformable neural radiance fields.
\newblock {\em ICCV}, 2021.

\bibitem{park2019few}
Seonwook Park, Shalini~De Mello, Pavlo Molchanov, Umar Iqbal, Otmar Hilliges,
  and Jan Kautz.
\newblock Few-shot adaptive gaze estimation.
\newblock In {\em Proceedings of the IEEE/CVF international conference on
  computer vision}, pages 9368--9377, 2019.

\bibitem{park2021talking}
Wooyeong Park, Jeongyun Heo, and Jiyoon Lee.
\newblock Talking through the eyes: User experience design for eye gaze
  redirection in live video conferencing.
\newblock In {\em International Conference on Human-Computer Interaction},
  pages 75--88. Springer, 2021.

\bibitem{vgg}
Karen Simonyan and Andrew Zisserman.
\newblock Very deep convolutional networks for large-scale image recognition.
\newblock In {\em ICLR}, 2015.

\bibitem{Simonyan15}
Karen Simonyan and Andrew Zisserman.
\newblock Very deep convolutional networks for large-scale image recognition.
\newblock In {\em International Conference on Learning Representations}, 2015.

\bibitem{sitzmann2019scene}
Vincent Sitzmann, Michael Zollh{\"o}fer, and Gordon Wetzstein.
\newblock Scene representation networks: Continuous 3d-structure-aware neural
  scene representations.
\newblock {\em Advances in Neural Information Processing Systems}, 32, 2019.

\bibitem{CAVE_0324}
B.A. Smith, Q. Yin, S.K. Feiner, and S.K. Nayar.
\newblock {G}aze {L}ocking: {P}assive {E}ye {C}ontact {D}etection for
  {H}uman?{O}bject {I}nteraction.
\newblock In {\em ACM Symposium on User Interface Software and Technology
  (UIST)}, pages 271--280, Oct 2013.

\bibitem{sugano2014learning}
Yusuke Sugano, Yasuyuki Matsushita, and Yoichi Sato.
\newblock Learning-by-synthesis for appearance-based 3d gaze estimation.
\newblock In {\em Proceedings of the IEEE conference on computer vision and
  pattern recognition}, pages 1821--1828, 2014.

\bibitem{vaswani2017attention}
Ashish Vaswani, Noam Shazeer, Niki Parmar, Jakob Uszkoreit, Llion Jones,
  Aidan~N Gomez, {\L}ukasz Kaiser, and Illia Polosukhin.
\newblock Attention is all you need.
\newblock {\em Advances in neural information processing systems}, 30, 2017.

\bibitem{wang2021facex}
Jun Wang, Yinglu Liu, Yibo Hu, Hailin Shi, and Tao Mei.
\newblock Facex-zoo: A pytorch toolbox for face recognition.
\newblock In {\em Proceedings of the 29th ACM International Conference on
  Multimedia}, pages 3779--3782, 2021.

\bibitem{wolf2010eye}
Lior Wolf, Ziv Freund, and Shai Avidan.
\newblock An eye for an eye: A single camera gaze-replacement method.
\newblock In {\em 2010 IEEE Computer Society Conference on Computer Vision and
  Pattern Recognition}, pages 817--824. IEEE, 2010.

\bibitem{gazedirector}
Erroll Wood, Tadas Baltrusaitis, Louis-Philippe Morency, Peter Robinson, and
  Andreas Bulling.
\newblock Gazedirector: Fully articulated eye gaze redirection in video, 2017.

\bibitem{xia2020controllable}
Weihao Xia, Yujiu Yang, Jing-Hao Xue, and Wensen Feng.
\newblock Controllable continuous gaze redirection.
\newblock In {\em Proceedings of the 28th ACM International Conference on
  Multimedia}, pages 1782--1790, 2020.

\bibitem{yu2019improving}
Yu Yu, Gang Liu, and Jean-Marc Odobez.
\newblock Improving few-shot user-specific gaze adaptation via gaze redirection
  synthesis.
\newblock In {\em Proceedings of the IEEE/CVF Conference on Computer Vision and
  Pattern Recognition}, pages 11937--11946, 2019.

\bibitem{yu2020unsupervised}
Yu Yu and Jean-Marc Odobez.
\newblock Unsupervised representation learning for gaze estimation.
\newblock In {\em Proceedings of the IEEE/CVF Conference on Computer Vision and
  Pattern Recognition}, pages 7314--7324, 2020.

\bibitem{zhang2022gazeonce}
Mingfang Zhang, Yunfei Liu, and Feng Lu.
\newblock Gazeonce: Real-time multi-person gaze estimation.
\newblock In {\em Proceedings of the IEEE/CVF Conference on Computer Vision and
  Pattern Recognition}, pages 4197--4206, 2022.

\bibitem{xgaze}
Xucong Zhang, Seonwook Park, Thabo Beeler, Derek Bradley, Siyu Tang, and Otmar
  Hilliges.
\newblock Eth-xgaze: A large scale dataset for gaze estimation under extreme
  head pose and gaze variation.
\newblock In {\em European Conference on Computer Vision (ECCV)}, 2020.

\bibitem{zhang18_etra}
Xucong Zhang, Yusuke Sugano, and Andreas Bulling.
\newblock Revisiting data normalization for appearance-based gaze estimation.
\newblock In {\em Proc. International Symposium on Eye Tracking Research and
  Applications (ETRA)}, pages 12:1--12:9, 2018.

\bibitem{deepappearence}
Xucong Zhang, Yusuke Sugano, Mario Fritz, and Andreas Bulling.
\newblock Appearance-based gaze estimation in the wild.
\newblock In {\em Proc. of the IEEE Conference on Computer Vision and Pattern
  Recognition (CVPR)}, pages 4511--4520, June 2015.

\bibitem{overface}
Xucong Zhang, Yusuke Sugano, Mario Fritz, and Andreas Bulling.
\newblock It’s written all over your face: Full-face appearance-based gaze
  estimation.
\newblock In {\em Computer Vision and Pattern Recognition Workshops (CVPRW),
  2017 IEEE Conference on}, pages 2299--2308. IEEE, 2017.

\bibitem{zhang2017mpiigaze}
Xucong Zhang, Yusuke Sugano, Mario Fritz, and Andreas Bulling.
\newblock Mpiigaze: Real-world dataset and deep appearance-based gaze
  estimation.
\newblock {\em IEEE transactions on pattern analysis and machine intelligence},
  41(1):162--175, 2017.

\bibitem{selflearning}
Yufeng Zheng, Seonwook Park, Xucong Zhang, Shalini~De Mello, and Otmar
  Hilliges.
\newblock Self-learning transformations for improving gaze and head
  redirection.
\newblock In {\em Neural Information Processing Systems (NeurIPS)}, 2020.

\bibitem{zhou2021pose}
Hang Zhou, Yasheng Sun, Wayne Wu, Chen~Change Loy, Xiaogang Wang, and Ziwei
  Liu.
\newblock Pose-controllable talking face generation by implicitly modularized
  audio-visual representation.
\newblock In {\em Proceedings of the IEEE/CVF conference on computer vision and
  pattern recognition}, pages 4176--4186, 2021.

\bibitem{zhuang2022mofanerf}
Yiyu Zhuang, Hao Zhu, Xusen Sun, and Xun Cao.
\newblock Mofanerf: Morphable facial neural radiance field.
\newblock In {\em European Conference on Computer Vision}, 2022.

\bibitem{faceparsingmodel}
zllrunning.
\newblock Using modified bisenet for face parsing in pytorch.
\newblock \url{https://github.com/zllrunning/face-parsing.PyTorch}.

\end{thebibliography}
}

\end{document}